\documentclass{article}

    \PassOptionsToPackage{numbers, compress}{natbib}

     \usepackage[preprint]{neurips_2019}

\usepackage[utf8]{inputenc} 
\usepackage[T1]{fontenc}    
\usepackage{hyperref}       
\usepackage{url}            
\usepackage{booktabs}       
\usepackage{amsfonts}       
\usepackage{nicefrac}       
\usepackage{microtype}      
\usepackage{pdfpages}

\usepackage{amsmath}
\usepackage{graphicx}
\usepackage{dsfont}
\usepackage{booktabs}
\usepackage{multirow}
\usepackage{wrapfig}
\usepackage{float}
\usepackage{subfigure}

\usepackage{color}
\definecolor{darkgreen}{rgb}{0,0.6,0}

\definecolor{darkblue}{rgb}{0,0,0.6}

\definecolor{darkred}{rgb}{0.6,0,0}

\title{Neural reparameterization improves\\
structural optimization
}

\author{%
  Stephan Hoyer \\
  Google Research\\
  \texttt{shoyer@google.com} \\
   \And
   Jascha Sohl-Dickstein \\
   Google Research \\
   \texttt{jaschasd@google.com} \\
   \And
   Sam Greydanus\\
   Google Research \\
   \texttt{sgrey@google.com} \\
}

\begin{document}

\maketitle

\begin{abstract}
  Structural optimization is a popular method for designing objects such as bridge trusses, airplane wings, and optical devices. Unfortunately, the quality of solutions depends heavily on how the problem is parameterized. In this paper, we propose using the implicit bias over functions induced by neural networks to improve the parameterization of structural optimization. Rather than directly optimizing densities on a grid, we instead optimize the parameters of a neural network which outputs those densities. This reparameterization leads to different and often better solutions.
  On a selection of 116 structural optimization tasks, our approach produces the best design 50\% more often than the best baseline method.
\end{abstract}

    \begin{wrapfigure}[14]{r}{0.51\textwidth}
    \centering
    \vspace{-1.5cm}%
    \hspace{0mm}%
    \includegraphics[width=0.5\textwidth, clip, trim=0mm 0mm 0mm 0mm]{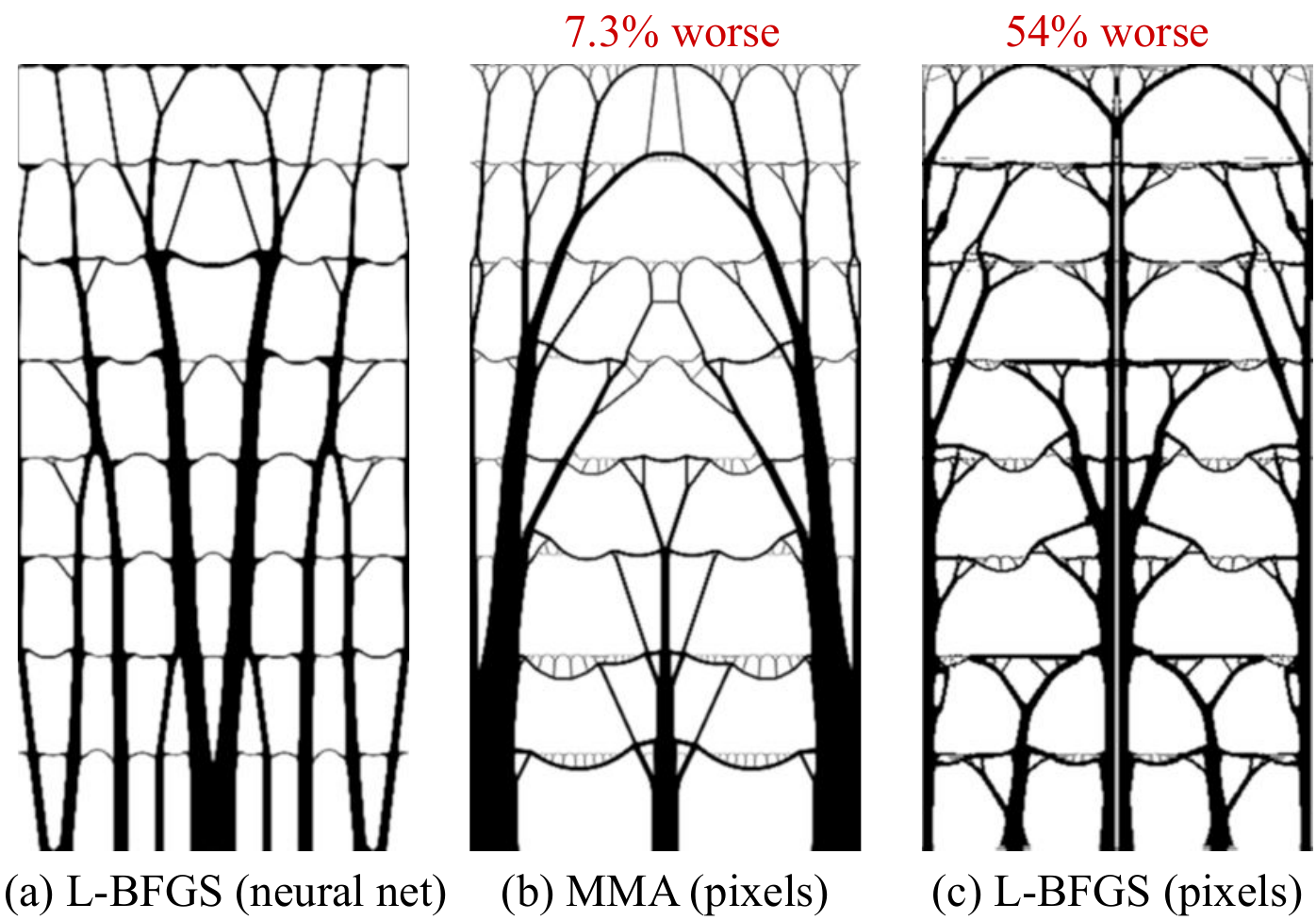}\\
    \vspace{-0.2cm}%
    \caption{A multi-story building task. Figure (a) is a structure optimized in CNN weight space. Figures (b) and (c) are structures optimized in pixel space.}
    \label{fig:multistory}
    \end{wrapfigure}

\section{Introduction}

One of the driving forces behind the success of deep computer vision models is the so-called ``deep image prior" 
of convolutional neural networks (CNNs). This phrase loosely describes a set of inductive biases, present even in untrained models, that make them effective for image processing. Researchers have taken advantage of this effect to perform inpainting, noise removal, and super-resolution on images with an untrained model \cite{ulyanov2018deep}.

There is growing evidence that this implicit prior extends to domains beyond natural images. Some examples include style transfer in fonts \cite{Azadi_2018_CVPR}, uncertainty estimation in fluid dynamics \cite{zhu2019physics}, and data upsampling in medical imaging \cite{dittmer2018regularization}. Indeed, whenever data contains translation invariance, spatial correlation, or multi-scale features, the deep image prior may be a useful tool.

One field where these characteristics are important -- and where the deep image prior is under-explored -- is computational science and engineering.
Here, parameterization is extremely important -- substituting one parameterization for another has a dramatic effect. Consider, for example, the task of designing a multi-story building via structural optimization. The goal is to distribute a certain quantity of building material over a two-dimensional grid in order to maximize the resilience of the structure. As Figure \ref{fig:multistory} shows, different optimization methods (LBFGS \citep{lbfgs} vs. MMA \citep{mma}) and parameterizations (pixels vs. neural net) have big consequences for the final design.

How can we harness the deep image prior to better solve problems in computational science? In this paper, we propose reparameterizing optimization problems from the basis of a grid to the basis of a neural network. We use this approach to solve 116 structural optimization tasks and obtain solutions that are quantitatively and qualitatively better than the baselines.

\section{Methods}

\begin{figure}[t!]
\centering
\includegraphics[width=0.95\textwidth]{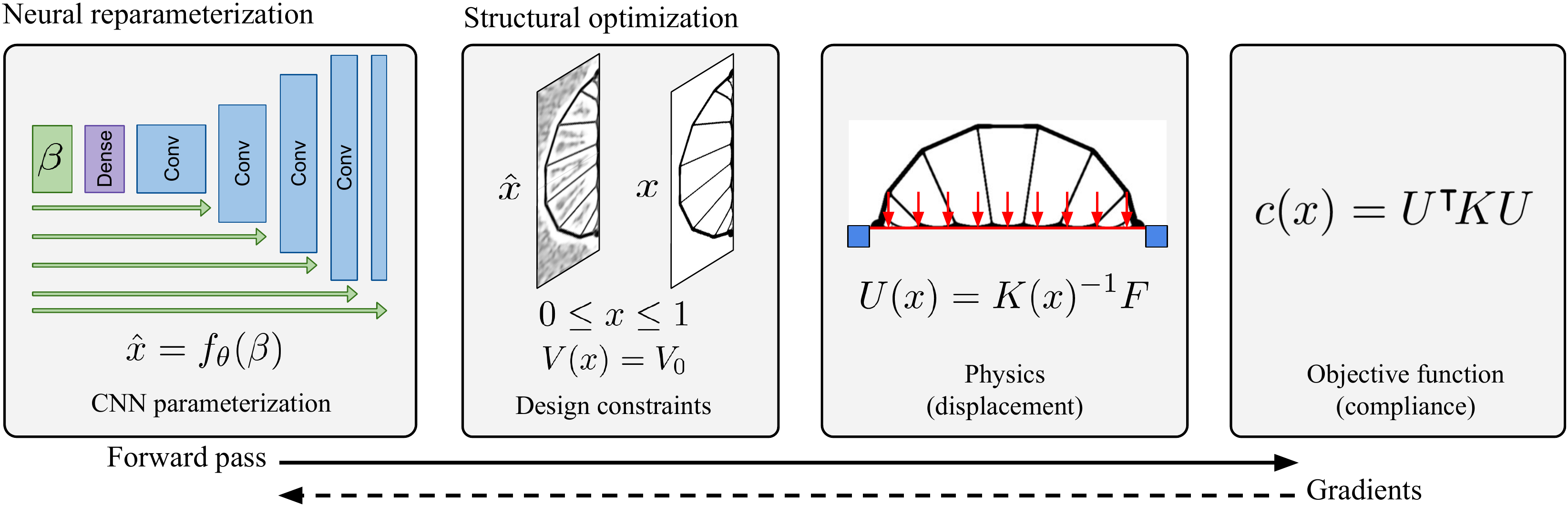}
\caption{Schema of our approach to reparameterizing a structural optimization problem with a neural network. Each of these steps -- the CNN parameterization, the constraint step, and the physics simulation -- is differentiable. We implement the forward pass as a TensorFlow graph and compute gradients via automatic differentiation.}
\label{fig:schema}
\end{figure}

While we apply our approach to structural optimization in this paper, we emphasize that it is generally applicable to a wide range of optimization problems in computational science.
The core strategy is to write the physics model in an automatic differentiation
package with support for neural networks, such as Jax, TensorFlow, or PyTorch.
We emphasize that the differentiable physics model need not be written from scratch:
adjoint models, as these are known in the physical sciences, are widely used \cite{Plessix2006-dw, Errico1997-to, Giles2000-np}, and software packages exist for computing them automatically \cite{Farrell2013-fg}.

The full computational graph begins with a neural network forward pass, proceeds to enforcing constraints and running the physics model, and ends with a scalar loss function (``compliance" in the context of structural optimization). 
Figure \ref{fig:schema} gives an overview of this process. Once we have created this graph, we can recover the original optimization problem by performing gradient descent on the inputs to the constraint step ($\hat x$ in Figure \ref{fig:schema}). Then we can reparameterize the problem by optimizing the weights and inputs ($\theta$ and $\beta$) of a neural network which outputs $\hat{x}$.

\textbf{Structural optimization.} 
We demonstrate our reparameterization approach on the domain of structural optimization. 
The goal of structural optimization is to use a physics simulation to design load-bearing structures, given constraints such as conservation of volume.
We focus on the general case of free-form design without configuration constraints, known as topology optimization \cite{topoopt_book}.

Following the ``modified SIMP" approach described by \cite{88lines}, we begin with a discretized domain of linear finite elements on a regular square grid. 
The physical density $\tilde x_{ij}$ at grid element (or pixel) $(i,j)$ is computed by applying a cone-filter with radius 2 on the input densities $x_{ij}$.
Then, letting $K(\tilde x)$ be the global stiffness matrix, $U(K,F)$ the displacement vector, $F$ the vector of applied forces, and $V(\tilde x)$ the total volume, we can write the optimization objective as:
\begin{align}
    \min_x: c(x) = U^T K U,
    \quad\text{such that:}\quad
    K U = F, \quad
    V(x) = V_0, \quad \text{and }
    0 \leq x_{ij} \leq 1\ \  \forall (i,j).
    \label{eq:topology-optimization}
\end{align}
We implemented this algorithm in NumPy, SciPy and Autograd \cite{maclaurinautograd}.
The computationally limiting step is the linear solve $U=K^{-1}F$, for which we use a sparse Cholesky factorization~\cite{cholmod}.

One key challenge was enforcing the volume and density constraints of Equation \eqref{eq:topology-optimization}.
Standard topology optimization methods satisfy these constraints directly, but only when directly optimizing the design variables $x$.
Our solution was to enforce the constraints in the forward pass, by mapping unconstrained logits $\hat x$ into valid densities $x$ with a constrained sigmoid transformation:
\begin{align}
    x_{ij} = 1/(1 + \exp[\hat x_{ij} - b(\hat x, V_0)]),
    \quad\text{such that:} \quad
    V(x) = V_0.
\end{align}
where $b(\hat x, V_0)$ is solved for via binary search on the volume constraint. In the backwards pass, we differentiate through the transformation at the optimal point using implicit differentiation \cite{griewank2002reduced}.

\textbf{A note on baselines.} Structural optimization problems are sensitive not only to choice of parameterization but also to choice of optimization algorithm. Unfortunately, standard topology optimization algorithms like the Method of Moving Asymptotes (MMA) \cite{mma} and the Optimality Criteria (OC) \cite{Bendsoe1995-ci} are ill-suited for training neural networks. How, then, can we separate the effect of parameterization from choice of optimizer? Our solution was to use a standard gradient-based optimizer, L-BFGS~\cite{Nocedal1980-yg}, to train both the neural network parameterization (CNN-LBFGS) and the pixel parameterization (Pixel-LBFGS).
We found L-BFGS to be significantly more effective than stochastic gradient descent when optimizing a single design, similar to findings for style transfer \cite{Gatys2016-wk}.

\begin{wrapfigure}[17]{r}{0.5\textwidth}
\centering
\vspace{-.5cm}%
\hspace{0mm}%
\includegraphics[width=0.5\textwidth]{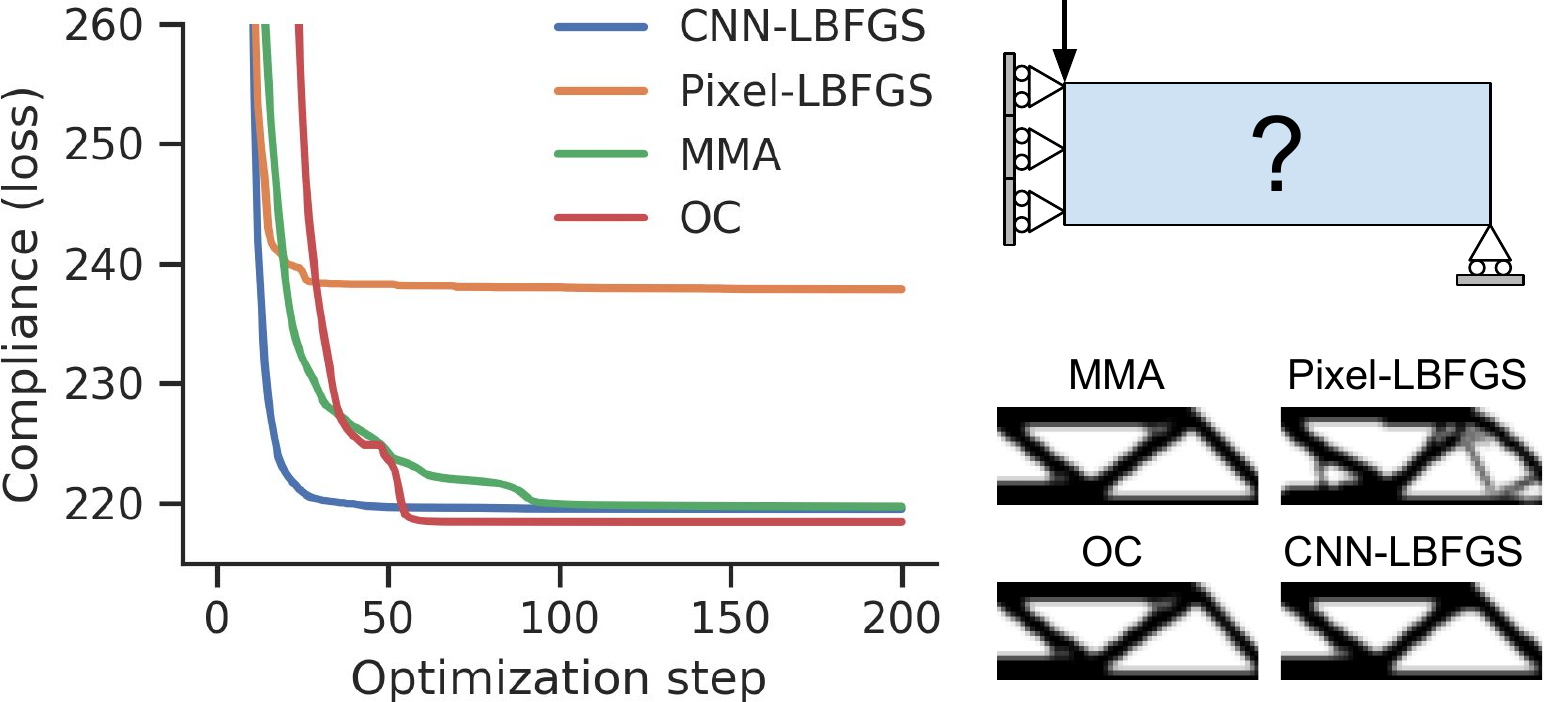}\\
\caption{Comparing baselines on the MBB beam example, on a $60 \times 20$ grid. Whereas Pixel-LBFGS and CNN-LBFGS use the same optimizer, we found that MMA and OC are much stronger baselines, so we decided to report all three.
We use the implementation of MMA from NLopt~\cite{nlopt}. We re-implemented OC, but verified the results agree exactly on the tasks reported in~\cite{88lines}.
}
\label{fig:mbb}
\end{wrapfigure}

Since constrained optimization is often much more effective at topology optimization (in pixel space, at least), we also report the MMA and OC results.
In practice, we found that these provided stronger baselines than Pixel-LBFGS. Figure \ref{fig:mbb} is a good example: it shows structural optimization of an MBB beam using the three baselines. All methods except Pixel-LBFGS converge to similar, near-optimal solutions.

\textbf{Choosing the 116 tasks.} In designing the 116 structural optimization tasks, our goal was to create a distribution of diverse, well-studied problems with real-world significance. We started with a selection of problems from \cite{topo_benchmarks} and \cite{Sokol2011-xs}. Most of these classic problems are simple beams with only a few forces, so we hand-designed additional tasks reflecting real-world designs including bridges with various support restrictions, trees, ramps, walls and buildings. The final tasks fall into 28 categories, with $V_0 \in [0.05,  0.5]$ and between $2^{11}$ to $2^{16}$ elements.

\textbf{Neural network methods.} Our convolutional neural network architecture was inspired by the U-net architecture used in the Deep Image Prior paper \cite{ulyanov2018deep}. We were only interested in the parameterization capabilities of the this model, so we used only the second, upsampling half of the model. We also made the first activation vector ($\beta$ in Figure \ref{fig:schema}) into a trainable parameter. Our model consisted of a dense layer into $32$ image channels, followed by five repetitions of tanh nonlinearity, $2$x bilinear resize (for the middle three layers), global normalization by subtracting the mean and dividing by the standard deviation, a 2D convolution layer, and a learned bias over all elements/channels. The convolutional layers used $5\times5$ kernels and had $128$, $64$, $32$, $16$, and $1$ channels respectively.

\section{Analysis}

\begin{wrapfigure}[17]{r}{0.5\textwidth}
\centering
\vspace{-1.7cm}%
\hspace{0mm}%
\includegraphics[width=0.5\textwidth]{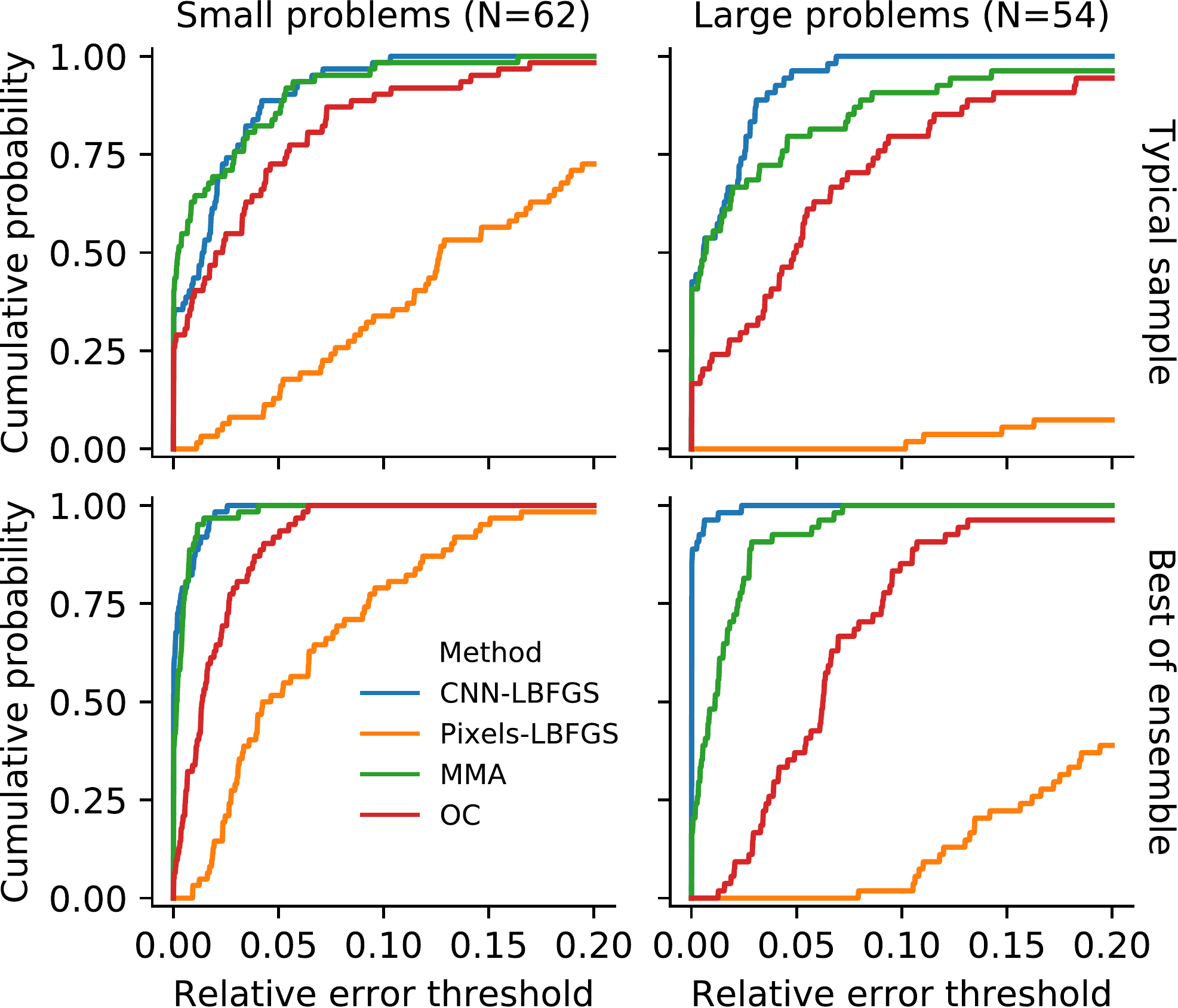}\\
\caption{Empirical distribution of the relative error across design tasks. The $x$-axes measure design error relative to the best overall design. The $y$-axes measure the probability that the method's solution has an error below the $x$-axis threshold.}
\label{fig:error}
\end{wrapfigure}

We found that reparameterizing structural optimization problems with a neural network gave equal performance to MMA on small problems and compellingly better performance on large problems. On both small and large problems, it produced much better designs than OC and Pixel-LBFGS.

For each task, we
report typical (median over 101 random seeds for the CNN, constant initialization for the other models\footnote{Constant initialization was better than the median for all baseline models.}) performance and ``best-of-ensemble" performance (with the same initializations for all models, taken from the untrained CNN). Figure \ref{fig:error} summarizes our results; its second column of plots show how on large problems (defined by $\geq 2^{15}$ grid points) the CNN-LBFGS solutions were more likely to have low error.

\begin{wrapfigure}[28]{r}{0.5\textwidth}
\centering
\vspace{-0.35cm}%
\hspace{0mm}%
\includegraphics[width=0.5\textwidth]{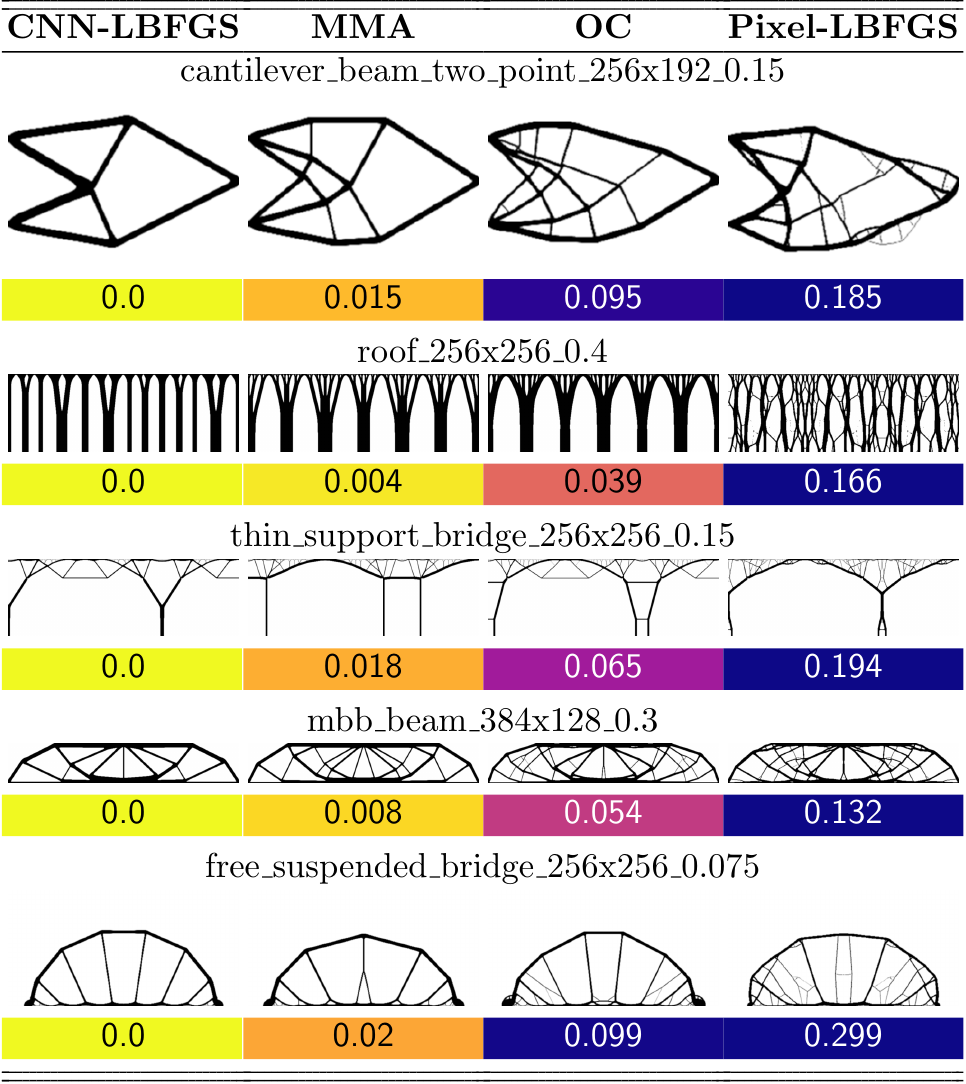}\\
\caption{Qualitative examples of structural optimization via reparameterization. The scores below each structure measure relative difference between the design and the best overall design in that row. The ``best of ensemble'' CNN-parameterized solutions were best or near-best (score $\leq$ 0.005) in 99 out of 116 tasks including these five, vs.\ 66 out of 116 tasks for MMA. The CNN solutions are qualitatively different from the baselines and often involve simpler and more effective structures.}
\label{fig:structs}
\end{wrapfigure}

\textbf{Why do large problems benefit more?}
Returning to the literature, we found that finite grids can suffer from a ``mesh-dependency problem", with varying solutions as grid resolution changes \cite{Sigmund1998NumericalII}.
When grid resolution is high, small-scale ``spiderweb" structures tend to form first and then interfere with the development of large-scale structures.
We suspected that optimizing the weights of a CNN allowed us to instead optimize structures at several spatial scales at once, thus improving optimization dynamics.
To investigate this idea, we plotted structures from all 116 design tasks (see ``Ancilliary files'' for this paper on arXiv.org). Then we chose five examples to highlight and showcase important qualitative trends (Figure \ref{fig:structs}).

\textbf{Reparameterized designs are often simpler.} 
The CNN-LBFGS designs have fewer ``spiderweb" artifacts as shown in the cantilever beam, MBB beam, and suspended bridge examples. On the cantilever beam, CNN-LBFGS used a total of eight supports whereas MMA used eighteen. We see simpler structures as evidence that the CNN biased optimization towards large-scale structure. This effect was particularly pronounced for large problems, which may explain why they benefited more.

\textbf{Convergence to different solutions.} We also noted that the baseline structures resembled each other more closely than they did CNN-LBFGS. In the thin support bridge example, the baseline designs feature double support columns whereas CNN-LBFGS used a single support with treelike branching patterns. In the roof task, the baselines use branching patterns, but the CNN-LBFGS uses pillars.

\section{Related work}

\textbf{Parameterizing topology optimization.} The most common parameterization for topology optimization is a grid mesh \cite{88lines, 99lines, zhu2016topology}. Sometimes, polyhedral meshes are used \cite{gain2015topology}. Some domain-specific structural optimizations feature locally-refined meshes and multiple load case adjustments \cite{krog2004topology}. Like locally refined meshes, our method permits structure optimization at multiple scales. Unlike them, our method permits optimization on both scales \textit{at once}.

\textbf{Neural networks and topology optimization.} Several papers have proposed replacing topology optimization methods with CNNs \cite{banga20183d, sosnovik2019neural, alter2018structural,jiang2018data}. Most of them begin by creating a dataset of structures via regular topology optimization and then training a model on the dataset. While doing so can reduce computation, it comes at the expense of relaxing physics and design constraints. More problematically, these models can only reproduce their training data. In contrast, our approach produces \textit{better} designs that also obey exact physics constraints. One recent work resembles ours in that they use adjoint gradients to train a CNN model \cite{Jiang2019-yz}. Their goal was to learn a joint, conditional model over a range of related tasks, which is different from our goal of reparameterizing a single structure.

\section{Conclusions}

Choice of parameterization has a powerful effect on solution quality for tasks such as structural optimization, where solutions must be computed by numerical optimization. Motivated by the observation that untrained deep image models have good inductive biases for many tasks, we reparameterized structural optimization tasks in terms of the output of a convolutional neural network (CNN).
Optimization then involved training the parameters of this CNN for each task. The resulting framework produced qualitatively and quantitatively better designs on a set of 116 tasks.

\bibliography{reparam}
\bibliographystyle{reparam}


\end{document}